%% file: main.tex
\documentclass{article}

\usepackage{xcolor}
\usepackage{wrapfig,lipsum,booktabs}

\usepackage{amsmath}
\usepackage{multirow}

\usepackage[preprint]{neurips_2025}

\usepackage[utf8]{inputenc} 
\usepackage[T1]{fontenc}    
\usepackage{hyperref}       
\usepackage{url}            
\usepackage{booktabs}       
\usepackage{amsfonts}       
\usepackage{nicefrac}       
\usepackage{microtype}      
\usepackage{xcolor}         
\usepackage{graphicx} 
\usepackage{float}

\title{MV-RAG: Retrieval Augmented Multiview Diffusion} 

\author{
    \textbf{Yosef Dayani} \quad \textbf{Omer Benishu} \quad \textbf{Sagie Benaim} \\
    \\
    \\
    Hebrew University of Jerusalem \\
}

\begin{document}

\maketitle

\input{figures/teaser}

\input{sections/01_abstract}
\input{sections/02_introduction}
\input{sections/03_related_work}
\input{sections/04_method}

\input{sections/05_experiments}

\input{sections/06_conclusion}

\clearpage

{
    \small
    \bibliographystyle{abbrv}
    \bibliography{main}
}

\input{sections/07_appendix}

\end{document}

%% file: figures/teaser.tex
\begin{figure*}[!h]
\centering
\makebox[\textwidth]{%
    \includegraphics[width=1.2\textwidth]{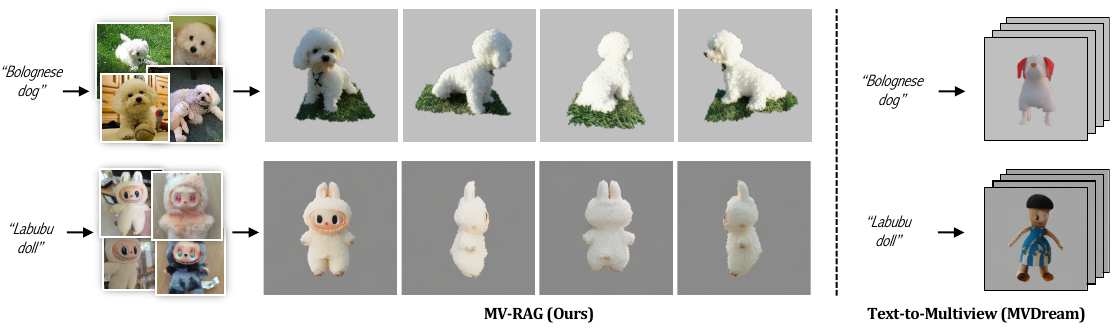}%
}
\caption{We introduce a retrieval-augmented diffusion framework for text-to-multiview generation. Given a text prompt, our method retrieves real-world images and adaptively leverages them together with the text, enabling faithful generation of out-of-distribution and newly emerging objects.}
\label{fig:teaser}
\vspace{-0.4cm}
\end{figure*}

%% file: sections/01_abstract.tex
\begin{abstract}
Text-to-3D generation approaches have advanced significantly by leveraging pretrained 2D diffusion priors, producing high-quality and 3D-consistent outputs. However, they often fail to produce out-of-domain (OOD) or rare concepts, yielding inconsistent or inaccurate results. To this end, we propose MV-RAG, a novel text-to-3D pipeline that first retrieves relevant 2D images from a large in-the-wild 2D database and then conditions a multiview diffusion model on these images to synthesize consistent and accurate multiview outputs. Training such a retrieval-conditioned model is achieved via a novel hybrid strategy bridging structured multiview data and diverse 2D image collections. This involves training on multiview data using augmented conditioning views that simulate retrieval variance for view-specific reconstruction, alongside training on sets of retrieved real-world 2D images using a distinctive held-out view prediction objective: the model predicts the held-out view from the other views to infer 3D consistency from 2D data. To facilitate a rigorous OOD evaluation, we introduce a new collection of challenging OOD prompts. Experiments against state-of-the-art text-to-3D, image-to-3D, and personalization baselines show that our approach significantly improves 3D consistency, photorealism, and text adherence for OOD/rare concepts, while maintaining competitive performance on standard benchmarks. Project page: \url{https://yosefdayani.github.io/MV-RAG/}
\end{abstract}

%% file: sections/02_introduction.tex
\section{Introduction}
\label{sec:introduction}

The automated generation of 3D content from textual descriptions is a task of great practical importance for various downstream applications, including game modeling~\cite{Gregory_2018, Lewis_Jacobson_2002}, computer animation~\cite{Parent_2012, Lasseter_1987}, and virtual reality~\cite{Schuemie_etal_2001}.

Current state-of-the-art approaches predominantly leverage powerful, pre-trained 2D text-to-image diffusion models~\cite{Song_etal_2020b_ScoreSDE, Ho_etal_2020_DDPM} as strong visual and semantic priors, either by optimization~\cite{Poole_etal_2022_DreamFusion, Lin_etal_2023_Magic3D, Liang_etal_2023_LucidDreamer} or by using them to train generative models that synthesize consistent multiview images~\cite{Liu_etal_2023_Zero123, Shi_etal_2023_MVDream, Long_etal_2024_Wonder3D}. Indeed, such approaches result in high-quality outputs for a diverse set of objects and scenes. 

However, they often falter when confronted with text prompts describing out-of-domain (OOD) or rare entities; 
in these settings, they often yield geometrically inconsistent results (e.g., poorly rendered unseen regions) or fail to adhere to the text, hallucinating details or substituting rare concepts with common ones.

A dominant text-to-3D approach leverages 2D priors via optimization, such as Score Distillation Sampling
(SDS)~\cite{Poole_etal_2022_DreamFusion, Lin_etal_2023_Magic3D, Liang_etal_2023_LucidDreamer}. These methods typically optimize a 3D representation, such as a NeRF~\cite{mildenhall2020nerfrepresentingscenesneural}, by distilling knowledge from pre-trained 2D text-to-image models. While producing high-fidelity results, SDS-based methods often struggle with geometric consistency and often inherit the 2D prior's failures on OOD/rare prompts, yielding flawed 3D assets.

To mitigate such issues, recent efforts~\cite{Seo_etal_2024_ReDream, chen2024sculpt3d} have explored retrieval augmentation within this optimization paradigm by retrieving existing 3D assets to provide explicit geometric priors during the generation. 
This enhances consistency for concepts in the 3D retrieval database but is limited by its scarce scale and diversity.

Another avenue involves 3D personalization techniques like DreamBooth3D~\cite{Raj_etal_2023_DreamBooth3D}, which first adapts a pretrained 2D text-to-image model to a specific subject using a few (3-6) user-provided images, and then applies SDS-based optimization. 
These methods excel at capturing subject-specific details but primarily operate via inference-time fine-tuning for each individual instance. Crucially, such lifting approaches still suffer from the inherent 3D geometry inconsistency and the optimization time of SDS-based approaches.

An alternative strategy to direct 3D optimization involves feed-forward approaches, particularly multi-view diffusion models~\cite{Shi_etal_2023_MVDream, Liu_etal_2023_Zero123, Long_etal_2024_Wonder3D}. These models aim to synthesize a consistent set of multi-view images from a given input (text or image), which can subsequently be employed for 3D reconstruction. Many such models are fine-tuned from general-purpose 2D diffusion backbones using large-scale 3D datasets such as Objaverse~\cite{deitke2023objaverse}, thereby enhancing their 3D awareness and enabling more stable and geometrically consistent generation compared to earlier 2D-lifting approaches.
Despite these advances, such models exhibit pronounced limitations when confronted with out-of-domain (OOD) or infrequent visual concepts. These challenges can be attributed to two main factors: (1) foundational 2D diffusion models, often used as the initialization point, may offer incomplete or biased representations of atypical concepts, and (2) the 3D datasets employed for fine-tuning, while extensive, often lack sufficient coverage, diversity, or geometric fidelity for uncommon entities. Consequently, these models tend to produce outputs with reduced photorealism, diminished multi-view consistency, or poor handling of unobserved regions when prompted with rare or novel concepts.

To this end, we develop a multiview diffusion model, MV-RAG, that, guided by an input text prompt, can also effectively condition its output on a set of relevant 2D images retrieved based on that text, ultimately producing a consistent set of multiview images that reflect the prompt's content. Our training strategy uniquely combines two sources of supervision: high-quality multiview images derived from structured 3D datasets, and diverse images sourced from large-scale 2D text-image collections. 
When training with multiview 3D data, we note that real-world images retrieved for a given text might look very different from the target multiview ground truth. 
To simulate retrieval variance, we augment ground truth multiview images (geometrically and semantically) creating diverse, "retrieval-like" conditioning views. 
Our model then learns to reconstruct the original scene, where the generation of each target view is guided explicitly by attending to visual tokens encoded from retrieved images. This allows for fine-grained conditioning based on these simulated retrieved inputs. To incorporate supervision from 2D data, we utilize $K+1$ images that are semantically similar to the input text from a given 2D text-image dataset. We then present $K$ of these images as conditional inputs, and our model is trained to generate the held-out image.  This novel objective pushes our model to infer 3D relationships and consistent appearances directly from sets of unstructured, real-world 2D views. Together, our hybrid training approach enables our model to learn robust geometric consistency from the multiview data, while simultaneously benefiting from the vast visual diversity and real-world appearance knowledge embedded in 2D image priors.

To adaptively balance the influence of the base model’s prior and the retrieved image signals, we introduce a fusion mechanism that dynamically adjusts their relative contribution based on how out-of-distribution the input prompt is.

We evaluate MV-RAG on OOD/rare concepts, where standard benchmarks fall short.
To this end, as current datasets focus on in-domain objects, we curated a new collection of 196 challenging prompts, corresponding to OOD/rare concepts, with corresponding retrieved images. Using this collection, we experimentally validate our approach against diverse state-of-the-art baselines. These include text-to-3D generation methods, image-to-3D techniques using our best retrieved images as input, and relevant 3D personalization approaches fine-tuned on our retrieved images. Our model demonstrates significant improvements in 3D consistency, photorealism, and text adherence for these challenging OOD/rare prompts, markedly outperforming existing methods, while maintaining competitive performance on standard in-domain benchmarks. Extensive ablation studies further validate the impact of our specific design choices. An illustration of our pipeline is provided in Fig.~\ref{fig:teaser}. 

\noindent \textbf{Contributions.} Our key contributions are: 
\textbf{(1)} Assessing the OOD problem in multiview generation and presenting a RAG pipeline to address this;
\textbf{(2)} Introducing a hybrid training scheme to bridge the gap between decoupled 2D retrievals and 3D objects;
\textbf{(3)} Developing the prior-guided attention mechanism to dynamically fuse the base model's prior with the external signals from retrievals;
\textbf{(4)} Introducing OOD-Eval, a new benchmark to facilitate further evaluation in this area.

%% file: sections/03_related_work.tex
\section{Related Work} 
\label{sec:related_work}

\input{figures/inference_illustration}

\noindent \textbf{3D Generation Using 2D Diffusion Models} \quad
Generating 3D content by leveraging strong priors from 2D diffusion models~\cite{Song_etal_2020b_ScoreSDE, Ho_etal_2020_DDPM} is a dominant paradigm. One major approach optimizes 3D representations, such as Neural Radiance Fields (NeRFs)~\cite{mildenhall2020nerfrepresentingscenesneural} and more recently 3D Gaussian Splatting (3DGS)~\cite{Kerbl_etal_2023_3DGS}, via Score Distillation Sampling (SDS)~\cite{Poole_etal_2022_DreamFusion, Lin_etal_2023_Magic3D, Liang_etal_2023_LucidDreamer, Tang_etal_2024_DreamGaussian, Yi_etal_2024_GaussianDreamer}, directly distilling knowledge from 2D priors. However, SDS often struggles with geometric consistency and fidelity due to weak 3D awareness in the priors~\cite{Hong_etal_2023_VSD, Shi_etal_2023_MVDream}.
Feed-forward multi-view diffusion models~\cite{Shi_etal_2023_MVDream, Liu_etal_2023_Zero123, Long_etal_2024_Wonder3D, MVAdapter_Huang_etal_2024}, often fine-tuning 2D diffusion priors with 3D dataset supervision, enhance geometric stability over 2D-lifting techniques by directly generating multiple consistent views~\cite{Shi_etal_2023_MVDream}. However, these models struggle with out-of-domain (OOD) or rare concepts due to limitations in 2D priors and insufficient 3D training data coverage, leading to reduced photorealism or inconsistent geometry. Related image-to-multiview approaches~\cite{Liu_etal_2023_Zero123, Shi_etal_2023_Zero123++, Wang_Shi_2023_ImageDream, Liu_etal_2023_SyncDreamer, Long_etal_2024_Wonder3D}, while effective with single clear inputs, are typically ill-suited for leveraging multiple, varied, unposed retrieved images. Our work builds on these advancements, targeting OOD generation by training a multiview diffusion model to effectively incorporate multiple retrieved 2D images.

\noindent \textbf{Retrieval Augmented Generation} \quad
Retrieval Augmented Generation (RAG) improves generative models by incorporating external information, aiding the handling of rare entities or specialized knowledge without full retraining, a successful paradigm in NLP~\cite{Lewis_etal_2020_RAG, Borgeaud_etal_2022_RETRO}. Notably, \cite{soudani2024fine} show that RAG is preferable to fine-tuning, especially for out-of-distribution (OOD) or less popular concepts, reinforcing the utility of retrieval-based methods in such settings. In 2D image synthesis, RAG methods similarly use retrieved images or text pairs to enhance fidelity for uncommon concepts or guide generation~\cite{Chen_etal_2022_ReImagen, Sheynin_etal_2022_KNNDiffusion, Blattmann_etal_2022_RetrievalAugDiffusion, Shalev-Arkushin_etal_2025_ImageRAG}. Recently, text-to-3D generation methods like RetDream~\cite{Seo_etal_2024_ReDream} and Sculpt3D~\cite{chen2024sculpt3d} retrieve existing \textit{3D assets} for geometric priors to improve optimization consistency. However, this is limited by the scarcity and diversity of 3D databases, especially for OOD or rare concepts. Inspired by findings in NLP, our work performs RAG in multiview generation via MV-RAG, leveraging abundant \textit{2D image datasets} to condition a multiview diffusion model, offering a more scalable way to ground OOD concept generation in real-world visual data.

\noindent \textbf{Personalization.} \quad
Personalization techniques adapt generative models to specific subjects using few examples. Originating in 2D text-to-image generation via methods such as embedding optimization or model fine-tuning~\cite{Gal_etal_2022_TextualInversion, Ruiz_etal_2023_DreamBooth}, these concepts extended to 3D. DreamBooth3D~\cite{Raj_etal_2023_DreamBooth3D}, for instance, adapts SDS optimization with a personalized 2D prior fine-tuned on few-shot examples. Similarly, multiview diffusion models like MVDream~\cite{Shi_etal_2023_MVDream} can be fine-tuned with DreamBooth-like principles for subject-specific generation with enhanced multiview consistency. These 3D personalization methods typically rely on inference-time optimization or fine-tuning per subject using a few examples. While designed for specific subjects, such approaches could be applied using retrieved image sets. Our work differs by integrating retrieved 2D image sets representing general concepts directly into our multiview diffusion model's main training phase, instead of inference-time adaptation with limited examples. This aims to improve the model's intrinsic ability to generate a broader range of concepts, especially OOD ones, without per-subject, inference-time adaptation.

%% file: figures/inference_illustration.tex
\begin{figure*}[t]
\includegraphics[width=0.95\linewidth]{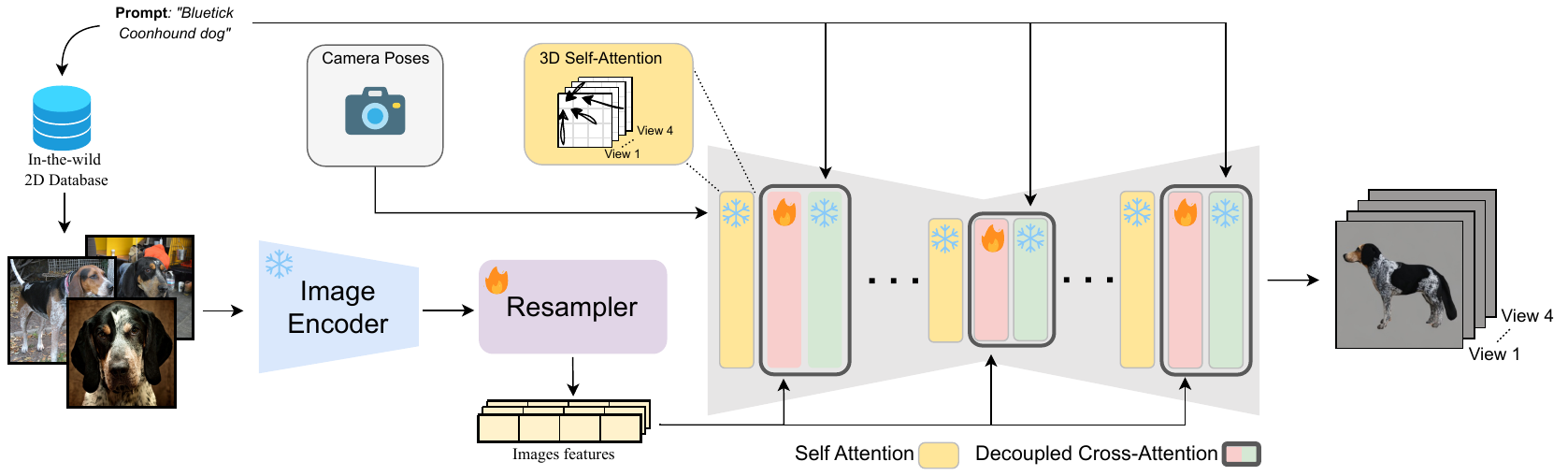}
\caption{\textbf{Overview of our pipeline.} Given a text prompt, we retrieve $k$ relevant images from an in-the-wild 2D image corpus. Local features are extracted from each image, projected through a Resampler and integrated into retrieval-attention modules to guide the multi-view generation process.}
\label{fig:inference_illustration}
\vspace{-0.4cm}
\end{figure*}

%% file: sections/04_method.tex
\section{Method}
\label{sec:method}

\input{figures/train_illustration}

Given an input text prompt $p$, we first retrieve relevant 2D images corresponding to $p$ from a corpus of 2D text-image pairs. Then, we use these images, along with $p$, to guide the generation process of a multi-view diffusion model. An overview of our approach is provided in Fig.~\ref{fig:inference_illustration}.
We begin by describing the training process, which involves a data preprocessing stage to prepare distinct conditioning and target data for 2D or 3D training modes, followed by the training of our multiview diffusion model.

\subsection{Training Data Preprocessing}
\label{sec:preprocessing}

Our model training leverages geometric grounding from 3D datasets and diversity from 2D datasets. For both, we prepare 2D conditioning images related to a text $p$, simulating inference-time retrieval, alongside target supervision data.
Fig.~\ref{fig:train_illustration} illustrates our separate 2D and 3D training modes. 

\noindent \textbf{2D Data Mode Supervision.} \quad
For this mode, we utilize a large-scale 2D text-image dataset, focusing on in-the-wild images that are neither posed nor aligned. 
For a text prompt $p_i$, we consider $K+1$ relevant images, from which we designate $K$ as the conditioning retrieved views $\mathcal{I}_{\text{ret}}=\{I_i\}^{K}_{i=1}$, and the remaining image, $I_{\text{target}}$, serves as the target image on which the diffusion loss is computed. This process yields training samples of the form $\mathcal{D}_{2D}=\{p, \mathcal{I}_{\text{ret}}, I_{\text{target}}\}$. No ground truth camera poses are assumed for these images.

\noindent \textbf{3D Data Mode Supervision.} \quad
For 3D data, we assume a dataset comprising text prompts and corresponding 3D object models. For each 3D object, we render a set of $N$ ground truth target views $I_{\text{target}}=\{I_i\}^{N}_{i=1}$ at $N$ camera poses $\mathcal{C}$. We follow MVDream and use 4 orthogonal camera poses. To simulate the diverse nature of retrieved conditioning images, we render the object from $K$ additional random poses followed by a sequence of random augmentations. This yields conditioning views $\mathcal{I}_{\text{ret}}=\{I_i\}_{i=1}^{K}$. 
We apply a combination of geometric and semantic augmentations designed to mimic in-the-wild variability and enhance generalization. This yields training samples of the form $\mathcal{D}_{3D}=\{p,\mathcal{C}, \mathcal{I}_{\text{ret}},\mathcal{I_{\text{target}}}\}$. See Sec.\ref{sec:supp_implementation_details} for additional details.

\subsection{Retrieved and Augmented Image Encoding}
\label{sec:encoding}
A core component of our approach is the effective encoding of the $K$ conditioning images into sequences of conditioning tokens. To achieve this, we employ a pre-trained image encoder, $E$, specifically the Vision Transformer (ViT) version of  CLIP~\cite{Radford_etal_2021_CLIP}, to extract rich features from each conditioning image $I_i$. We extract the sequence of patch-level (local) features from the penultimate layer of the transformer, $F_i = E(I_i)$, 
for a richer, spatially descriptive representation over a global embedding. $E$ is frozen and not fine-tuned. 

To efficiently incorporate the rich spatial information from each retrieved image, we apply a learnable Resampler module $\Theta_R$ inspired by the Perceiver Resampler architecture~\cite{Jaegle_etal_2021_Perceiver} and its use in IP-Adapter variants~\cite{Ye_etal_2023_IPAdapter}. This module distills the salient visual information from $F_i$ into a compact set $N_t$ of tokens $T_i = \Theta_R(F_i)$, using a small set of learnable queries attending to $F_i$. Following prior work, we set $N_t=16$. This enables effective conditioning of the diffusion model while reducing computational overhead. The resulting token sequences are then used for conditioning the cross-attention layers, as described next.

\subsection{Retrieval-Conditioned Multiview Diffusion}
\label{sec:model_conditioning_training}

The encoded tokens are then fed into a multiview diffusion model, based on MVDream~\cite{Shi_etal_2023_MVDream}, which extends a 2D text-to-image U-Net architecture for multiview generation.
Following MVDream, we incorporate camera pose embeddings for geometric guidance and modify the U-Net's self-attention layers. These layers are inflated to operate jointly over features from all generated views, forming a 3D-aware self-attention mechanism that promotes multiview consistency.

While MVDream relies solely on text-based cross-attention, we replace this mechanism with a decoupled cross-attention module that incorporates encoded tokens from both the text prompt and the retrieved images.
Specifically, the tokens from the conditioning images are processed by a dedicated, trainable cross-attention branch, yielding retrieval-guided features denoted as $f_{\text{ret}}$. 

For this cross-attention branch, we follow the design of IP-Adapter~\cite{Ye_etal_2023_IPAdapter}, where the U-Net query features $Q_i$ (generated via a shared query projection $\theta_Q$) attend separately to keys and values from the retrieved tokens $T_i$ and the text embedding. The retrieved tokens are processed through learnable projections $\theta_{K_{\text{ret}}}$ and $\theta_{V_{\text{ret}}}$ to produce $f_{\text{ret}}$, while the text embedding is processed through frozen projections $\theta_{K_{\text{txt}}}$ and $\theta_{V_{\text{txt}}}$, inherited from a pretrained MVDream model, yielding $f_{\text{txt}}$. We note that the shared query projection $\theta_Q$ is also frozen. 

This results in a decoupled cross-attention mechanism. During training, we integrate the text and retrieval features as $f = \lambda f_{\text{txt}} + f_{\text{ret}}$, where $\lambda$ is a hyperparameter. Empirically, we find that small values of $\lambda$ ease the adaptation of the newly introduced retrieval branch during training.

\subsection{2D and 3D Training Modes}

Our full architecture then comprises the encoder, resampler, and U-net-based architecture (comprising of self-attention and decoupled cross-attention layers).  This architecture is trained jointly using the two data modes described below:

\noindent \textbf{3D Data Mode: Multiview Reconstruction.} \quad
When training with samples derived from 3D assets, the model reconstructs a set of predicted views given their camera poses $\mathcal{C}$. The U-Net's self-attention layers operate across all $N$ view latents, enforcing cross-view consistency. Each target view is conditioned on its camera pose, and the text prompt $p$ provides global guidance via its features $f_{\text{txt}}$. The visual tokens aggregated from all $K$ augmented conditioning images $\mathcal{I}_{\text{ret}}$ are used to compute the retrieval attention features $f_{\text{ret}}$, jointly guiding the generation of all $N$ target views. A multiview reconstruction loss, $\mathcal{L}_{MV}(\theta,p, \mathcal{C}, \mathcal{I}_{\text{ret}},\mathcal{I}_{\text{pred}})$, is applied across all target views. 

\noindent \textbf{2D Data Mode: Held-out View Prediction.} \quad
When training with samples from 2D datasets, the objective is to predict the single held-out image $I_{\text{target}}$ based on the text prompt $p$ and tokens from the $K$ conditioning retrieved images $\mathcal{I}_{\text{ret}}$. In this scenario, as only a single target view is generated, the U-Net's self-attention layers inherently function as standard 2D self-attention, operating within that single view's features. The text prompt $p$ and the tokens from the retrieved images provide conditioning via $f_{\text{txt}}$ and $f_{\text{ret}}$ respectively. Crucially, consistent with MVDream's training on 2D data, no explicit camera pose information is provided. 

\subsection{Inference Process}
\label{sec:inference}
At inference time, given an input text prompt $p$, we first retrieve the top $K$ relevant 2D images $\mathcal{I}_{\text{ret}}$ from our diverse 2D database using the BM25-based text similarity approach. 
To improve relevance, we compute prompt-caption similarity and discard images below a threshold, yielding \( K' \leq K \) images. If no images pass the threshold, we disable retrieval-attention and fall back to the base model. These $K$ images are then encoded into visual tokens as detailed in Section~\ref{sec:encoding}. Our trained multiview diffusion model then generates $N$ consistent views conditioned on the text prompt $p$ and the set of retrieved tokens, utilizing specified camera poses for the target views. 

\input{figures/prior_guided_figure}

\noindent \textbf{Prior-guided attention.} \quad
We introduce an adaptive fusion coefficient $\alpha$ that dynamically balances the influence of the model's prior knowledge and the retrieved external signals, based on how out-of-distribution (OOD) a prompt is. Diffusion models learn to approximate the score function $\nabla_x \log p(x|y)$, the gradient of the log data density which by definition points toward higher-probability regions of the data distribution~\cite{song2020generativemodelingestimatinggradients,ho2020denoisingdiffusionprobabilisticmodels}. Thus, if a concept is in-domain, the base model’s score will guide denoising toward an accurate reconstruction, whereas for OOD concepts the reconstruction will deviate.

To estimate $\alpha$ during inference, without ground-truth multiview, we first perform a short forward pass using only the base model’s text-based attention $f_{\text{txt}}$ (with the retrieval module disabled) for 10 DDIM steps, generating an initial candidate output. We then measure its similarity to the retrieved images using DINOv2 similarity~\cite{oquab2023dinov2}, which serves as a proxy for the base model’s confidence in capturing the concept: high similarity indicates in-domain content, so $\alpha$ favors $f_{\text{txt}}$; low similarity suggests OOD, shifting weight toward the retrieval-based attention $f_{\text{ret}}$. The two sources are fused adaptively as:
\[
f = \alpha \cdot f_{\text{txt}} + (\lambda' - \alpha) \cdot f_{\text{ret}},
\]
for a hyperparameter $\lambda'$, replacing the $f$ calculation used in training. The full model is then run with the retrieval module enabled to generate the final outputs. Fig.~\ref{fig:prior_guided} illustrates its effect.

%% file: figures/train_illustration.tex
\begin{figure*}[t]
\includegraphics[width=0.95\linewidth]{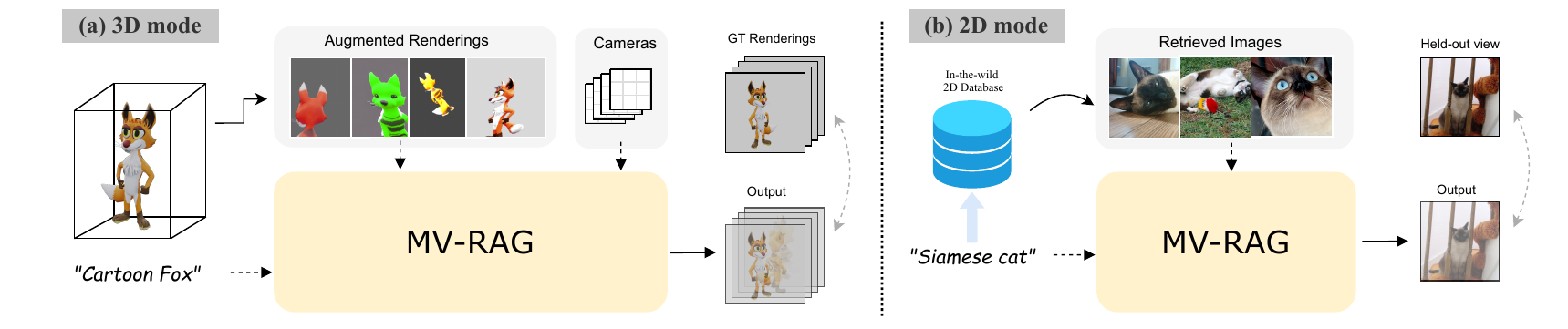}
\caption{\textbf{Overview of our training scheme.} Our model adopts a hybrid training strategy that alternates between two modes. \textbf{3D mode (left):} A 3D object is rendered to produce ground-truth multi-view images. Additional views are generated and subjected to heavy augmentations to serve as retrievals. These augmented views, along with the target camera parameters and the associated prompt, are provided as input to the model. \textbf{2D mode (right):} We retrieve $K+1$ images from a 2D training corpus, where $K$ images are used as retrievals and one held-out image serves as the target view. In this mode, the model performs 2D self-attention rather than 3D attention, and no target camera parameters are provided.
}
\label{fig:train_illustration}
\vspace{-0.5cm}
\end{figure*}

%% file: figures/prior_guided_figure.tex
\begin{figure}[t]
  \centering
  \includegraphics[width=\linewidth]{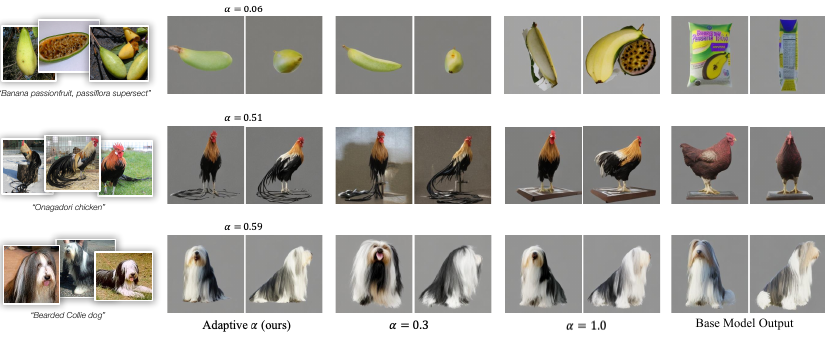}
  \caption{\textbf{Illustration of Prior-Guided Attention.}
The base model’s activations are leveraged proportionally to its prior knowledge of the object, controlled by the adaptive parameter $\alpha$. For an object where the base model has limited prior knowledge, a lower adaptive $\alpha$ reduces reliance on the base model. For an object well-represented by the base model, insufficient weighting (e.g., fixed $\alpha=0.3$) degrades results.} 
  \label{fig:prior_guided}
\end{figure}

%% file: sections/05_experiments.tex
\section{Experiments}
\label{sec:experiments}
\input{tables/main_comparison}
We evaluate our approach against relevant state-of-the-art baselines, as detailed below, both quantitatively and qualitatively, considering both OOD/rare and in-domain concepts. 

\noindent \textbf{Benchmarks} \quad
As current benchmarks lack OOD/rare concept coverage, we curated 196 examples from Wikipedia Commons~\cite{wikicommons} (not used in training).

Each example consists of a text prompt and multiple 2D retrieved images of the same concept. Importantly, texts were chosen to be far from any text (or concepts) seen during training. We call this evaluation set ``OOD-Eval". See Sec.\ref{sec:benchmarks} for additional details and examples. 
We also consider in-distribution objects, demonstrating that our success in OOD concepts is not compensated by worse in-domain results. To this end we consider a curated set of 50 in-domain objects from Objaverse-XL ~\cite{deitke2023objaversexluniverse10m3d}. For retrieval, we consider 2D images from the LAION-400M dataset~\cite{schuhmann2021laion}. For each text, we retrieve four 2D images. Unlike for OOD-Eval, we also have corresponding ground truth multiview images. We call this evaluation set ``IND-Eval". 

\noindent \textbf{Baselines} \quad
We compare MV-RAG against three categories of state-of-the-art methods.
(1) \textit{Text-to-multiview} generation: MVDream~\cite{Shi_etal_2023_MVDream} (closest in architecture to ours), MV-Adapter~\cite{MVAdapter_Huang_etal_2024} (text-conditioned), SPAD~\cite{kant2024spad}, and TRELLIS~\cite{xiang2025structured3dlatentsscalable} (text-conditioned).
(2) \textit{Image-to-multiview} generation applied to the retrieved 2D views: ImageDream~\cite{Wang_Shi_2023_ImageDream}, MV-Adapter~\cite{MVAdapter_Huang_etal_2024} (image-conditioned), Era3D~\cite{li2024era3d}, and TRELLIS~\cite{xiang2025structured3dlatentsscalable} (image-conditioned).
(3) \textit{3D personalization} models: we adopt MVDream’s optimization-based personalization approach~\cite{Shi_etal_2023_MVDream, Ruiz_etal_2023_DreamBooth}, applied to all $k$ retrieved views.
Further details are provided in Sec.\ref{sec:supp_baselines}.
\input{figures/2d_3d_ablation}

\subsection{Quantitative Evaluation}
\label{sec:quantative_eval}

\noindent \textbf{Metrics} \quad 
We assess generation quality with Inception Score (IS)~\cite{barratt2018noteinceptionscore} and FID~\cite{frechet_distance_fid} on the output poses.
To assess alignment to the input text, a natural choice would be to consider the CLIP~\cite{Radford_etal_2021_CLIP} similarity between the input text and the output multiview images produced by the model. 
However, we found that CLIP~\cite{Radford_etal_2021_CLIP} (specifically in image-text similarity) is unable to score rare/OOD concepts well, often assigning a low score for such text-image pairs. See Figure.~\ref{fig:clip_issues} and ~\ref{sec:supp_metrics} for further discussion and illustration. This is also demonstrated in \cite{zhuprompt}. 
As such, we evaluate image-image similarity between the generated views and held-out ground truth retrieved examples from our evaluation benchmark. Specifically, we compute the average similarity using CLIP~\cite{Radford_etal_2021_CLIP} and DINOv2~\cite{oquab2023dinov2}, both widely used for image representation. Additionally, we employ an \textit{Instance Retrieval (IR)} model~\cite{shao20221st} specifically trained to embed images of the same object instance close together in feature space, making it a more suitable choice for assessing entity-level visual alignment. 

\input{tables/user_study}

To evaluate 3D consistency, we adopt the procedure of~\cite{Wang_Shi_2023_ImageDream}, measuring how well a reconstruction model aligns with our generated views. Specifically, we sample four views and train a reconstruction model~\cite{tang2024lgm} on these “training views.” The model then predicts novel views (re-rendered in Fig.~\ref{tab:full-results}), whose fidelity and alignment with the training views are assessed using the metrics described above. Inconsistent multiview generations are expected to degrade reconstruction quality, leading to lower fidelity and alignment scores. Further details are provided in Sec.\ref{sec:supp_user_study}.

\textbf{User study.} \quad
To complement the quantitative metrics, we conduct a user study on prompts from the OOD-Eval dataset. For each prompt, we present four generated views from our model and the baselines, and ask participants to evaluate:
(Q1) \textit{Realism} — How realistic are the generated views?
(Q2) \textit{Alignment} — How well do the views match the input text?
(Q3) \textit{3D Consistency} — How consistent are the views across different perspectives?
For each question, outputs from all methods are shown in random order, and participants rank them on a scale of 1–5. Additional details are provided in Sec.~\ref{sec:supp_user_study}.
\input{tables/ind_compare_full}

\noindent \textbf{Evaluation on OOD/rare concepts} \quad
As shown in Tab.\ref{tab:full-results}, MV-RAG achieves strong performance across both evaluation modes.
In the \textit{4-views} setting, it outperforms all baselines on CLIP, DINO, and FID, while ranking second on IR (behind MV-Adapter (IM)) and IS (behind Era3D).
In the more challenging \textit{rerendered} setting which also reflects 3D consistency, MV-RAG leads on CLIP, DINO, IR, and FID, with Era3D attaining a higher IS.
Notably, MVDream and ImageDream, which share similar architectures but lack retrieval, consistently underperform across metrics.
The user study results in Tab.\ref{tab:user_study} further corroborate MV-RAG’s advantage, showing clear gains in realism, text alignment, and 3D consistency.

\noindent \textbf{Evaluation on in-domain concepts} \quad
We evaluate MV-RAG against all methods on the IND-Eval benchmark, which contains objects from Objaverse~\cite{deitke2023objaversexluniverse10m3d} dataset which is used for training in all baselines.
Reconstruction quality is measured using PSNR, SSIM, and LPIPS with respect to the ground-truth views in IND-Eval, while text-image alignment is assessed via CLIP~\cite{Radford_etal_2021_CLIP} and SigLIP~\cite{zhai2023sigmoid} similarity between the generated outputs and the input prompt.
As shown in Tab.~\ref{tab:ind_compare_full}, MV-RAG achieves results that are on par with, or slightly surpass, those of the baselines.

\input{figures/ablation_k}
\input{figures/qualitative_results}

\newpage
\subsection{Qualitative Evaluation}
In Fig.\ref{fig:qualitative}, we compare MV-RAG qualitatively against all baselines. Additional comparisons are provided in the appendix section (Figs.\ref{fig:comp_res}, \ref{fig:comp_res2}).
Text-to-3D models, which rely solely on text prompts, struggle with out-of-distribution (OOD) object structures due to the lack of visual priors for rare instances, often failing to capture key attributes or correct geometry.
Single-reference image-to-3D methods are limited by their single viewpoint, constraining pose diversity and missing occluded regions. Variations in lighting, texture, and visibility further degrade consistency and realism, restricting plausible reconstructions to views near the input perspective.
Although MVDreamBooth leverages multiple reference images, it fails to adequately handle variation across them (e.g., differing car colors) and struggles to construct accurate 3D structures, highlighting the challenge of integrating diverse visual cues effectively.

MV-RAG addresses these by leveraging multiple unposed images from a large 2D corpus, providing complementary viewpoints that enrich generation with diverse, relevant visual cues.
Crucially, our framework is designed to isolate view-invariant attributes such as object identity and disentangle nuisance factors like illumination, occlusion, and background clutter. As can be seen in Fig.~\ref{fig:qualitative}, this naturally results in more diverse and detailed multi-view outputs.

\noindent \textbf{Diversity and Utility. } \quad MV-RAG  enables diverse outputs, by using a different seed for a given input text and set of retrieved views. This is shown in Fig.~\ref{fig:diversity} (a).  In Fig.~\ref{fig:diversity}(b) we also demonstrate the MV-RAG utilizes different parts from retrieved views to generate target views  

\subsection{Ablation Study}

\noindent \textbf{Hybrid training} \quad
In Fig.~\ref{fig:2d_3d_ablation}, we present a qualitative ablation of our 2D mode, 3D mode, and augmentations. Without the 2D mode, the model struggles to separate the object from its in-the-wild background, leading to artifacts (e.g., a floating leash on a dog, a goat merged with a rock). Without the 3D mode, it fails to consistently distribute visual features across views, producing inaccurate shapes (e.g., tail or horn) and background inconsistencies. Removing augmentations still allows handling in-the-wild settings through the 2D mode but reduces robustness to the high variance in retrieved images, often yielding incorrect 3D structures.

\input{tables/retrieval_k}

\noindent \textbf{Number of Retrieved Images} \quad 
Fig.~\ref{fig:ablation_k}, considers the effect of number of retrieved images on  alignment/fidelity. As seen, using 4 views provides best overall performance. 

\noindent \textbf{Retrieval Method} \quad 
In Table~\ref{tab:retrieval_k}, we ablate the retrieval method used to obtain visual exemplars for conditioning. We evaluate retrieval performance using the OOD-Eval text prompts, where the goal is to retrieve the correct image-text pair from a combined collection of OOD-Eval and MS-COCO~\cite{lin2014microsoft}. Specifically, we compare four methods: CLIP~\cite{Radford_etal_2021_CLIP} with text-to-image similarity (TX-IM) and text-to-text similarity (TX-TX), SigLIP~\cite{zhai2023sigmoid} with TX-IM similarity, and BM25~\cite{Robertson1994OkapiAT}, a non-semantic bag-of-words model.

We observe that semantic retrieval models like CLIP and SigLIP struggle with rare concepts due to limited exposure and weak conceptual grounding. As shown in Figure~\ref{fig:clip_issues}, CLIP often assigns high similarity scores to incorrect generations or generic category matches, failing to distinguish nuanced, rare object names from broad class labels (see Table~\ref{tab:clip-retrieval-scores}). This issue directly carries over to retrieval: when queried with OOD descriptions, semantic models frequently return irrelevant results due to their inability to treat specific rare terms as meaningful. In contrast, BM25 relies purely on lexical overlap, matching keywords without relying on learned semantic priors. This makes it more robust in OOD scenarios where the semantics are underrepresented or misaligned in vision-language models.

%% file: tables/main_comparison.tex
\begin{table*}[t]
\centering
\caption{\textbf{Quantitative evaluation on OOD/rare concepts.} The models' performance is assessed on four orthogonal views. Multiview consistency is evaluated by reconstructing the 3D object from the multiview images. See Sec.~\ref{sec:experiments} for further details.}
\small
\renewcommand{\arraystretch}{1.1}
\resizebox{\textwidth}{!}{
\begin{tabular}{l|ccccc|ccccc}
\toprule
& \multicolumn{5}{c|}{\textbf{4-Views}} & \multicolumn{5}{c}{\textbf{Re-rendered}} \\
\textbf{Method} & CLIP ↑ & DINOv2 ↑ & IR ↑ & FID ↓ & IS ↑ & CLIP ↑ & DINOv2 ↑ & IR ↑ & FID ↓ & IS ↑ \\
\midrule
\multicolumn{11}{l}{\textbf{Text-to-3D}} \\
MVDream~\cite{Shi_etal_2023_MVDream} & 66.47 & 33.12 & 58.01 & 76.71 & 10.62 & 70.83 & 28.66 & 58.98 & 96.29 & 11.39 \\
MV-Adapter~\cite{MVAdapter_Huang_etal_2024} (TX) & 66.48 & 28.53 & 58.42 & 84.28 & 9.55 & 71.33 & 24.30 & 56.14 & 106.66 & 11.23 \\
SPAD~\cite{kant2024spad} & 65.23 & 19.39 & 48.54 & 167.49 & 9.18 & 64.46 & 12.29 & 43.80 & 176.66 & 8.90 \\
TRELLIS~\cite{xiang2025structured3dlatentsscalable} (TX) & 67.96 & 21.11 & 51.01 & 160.93 & 6.90 & 67.16  & 16.87 & 51.82 & 154.43  & 8.09  \\
\midrule
\multicolumn{11}{l}{\textbf{Image-to-3D}} \\
ImageDream-P~\cite{Wang_Shi_2023_ImageDream} & 69.20 & 45.01 & 65.64 & 68.40 & 12.11 & 70.44 & 32.77 & 60.17 & 103.24 & 12.84 \\
ImageDream-L~\cite{Wang_Shi_2023_ImageDream} & 67.55 & 39.48 & 63.93 & 84.69 & 9.45 & 70.16 & 29.60 & 58.66 & 120.37 & 10.42 \\
MV-Adapter~\cite{MVAdapter_Huang_etal_2024}(IM) & 69.74 & 49.14 & \textbf{71.05} & 72.71 & 12.88 & 71.53 & 35.25 & 60.36 & 107.95 & 12.64 \\
Era3D~\cite{li2024era3d} & 69.13 & 42.41 & 64.42 & 92.68 & \textbf{15.26} & 71.00 & 35.65 & 60.81 & 93.97 & \textbf{14.45} \\
TRELLIS~\cite{xiang2025structured3dlatentsscalable} (IM)  & 70.31 & 35.24 & 59.32 & 167.61 & 11.38 & 67.86  & 24.43 & 52.23 & 146.82  & 10.72 \\
\midrule
\multicolumn{11}{l}{\textbf{3D Personalization}} \\
MVDreamBooth~\cite{Shi_etal_2023_MVDream} & 66.14 & 36.22 & 55.09 & 82.73 & 11.55 & 68.38 & 27.91 & 54.33 & 107.07 & 11.92 \\
\midrule
\textbf{MV-RAG (Ours)} & \textbf{71.77} & \textbf{50.19} & 67.41 & \textbf{54.79} & 13.20 & \textbf{74.28} & \textbf{39.61} & \textbf{66.59} & \textbf{80.54} & 12.33 \\
\bottomrule
\end{tabular}
}
\label{tab:full-results}
\vspace{-0.5cm}
\end{table*}

%% file: figures/2d_3d_ablation.tex
\begin{figure}[t]
  \centering
  \includegraphics[width=\textwidth, trim={0cm 0cm 0cm 0cm}, clip]{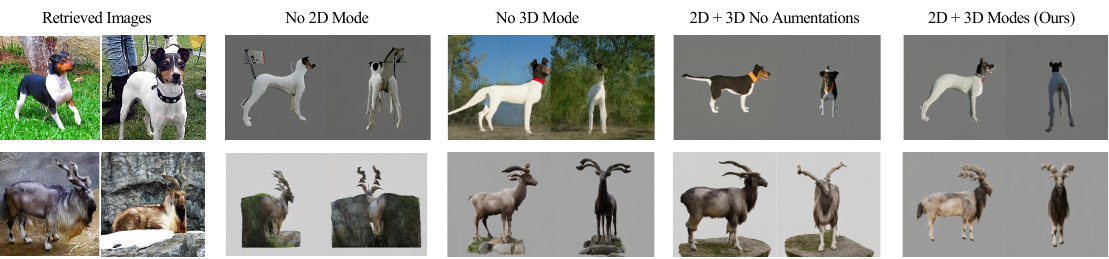}
  \caption{\textbf{Hybrid training ablations.} For the text prompts \textit{"Ratonero Bodeguero Andaluz dog"} (top) and \textit{"Markhor goat"} (bottom), we consider the output of our model when trained without our 2D/3D schemes in comparison to our full approach (shown on the RHS).}
  \label{fig:2d_3d_ablation}
\end{figure}

%% file: tables/user_study.tex
\begin{wraptable}{r}{5.6cm} 
\vspace{-0.2cm}
\centering
\caption{\textbf{User study.} A user study depicting (Q1-Realism), (Q2-Alignment) and (Q3-3D Consistency) reporting MOS (1-low, 5-high). }
{\small 
\setlength{\tabcolsep}{3pt} 
\begin{tabular}{lccc}
    \toprule
    & Q1 ↑ & Q2 ↑ & Q3 ↑ \\
    \midrule
    MVDream~\cite{Shi_etal_2023_MVDream} & 1.96 & 1.85 & 3.24 \\
    ImageDream-P~\cite{Wang_Shi_2023_ImageDream} & 2.25 & 2.6 & 3.03 \\
    MV-RAG (Ours) & \textbf{4.12} & \textbf{4.44} & \textbf{4.44}  \\
    \bottomrule
\end{tabular}
\setlength{\tabcolsep}{6pt} 
} 
\vspace{0.0cm}
\label{tab:user_study}
\end{wraptable}

%% file: tables/ind_compare_full.tex
\begin{wraptable}{r}{0.5\linewidth}
\vspace{-0.2cm}
\scriptsize 
\setlength{\tabcolsep}{3pt} 
\renewcommand{\arraystretch}{1.0} 
\centering
\caption{\textbf{Quantitative evaluation on in-domain concepts.}}
\begin{tabular}{lccccc}
\toprule
\textbf{Model} & \textbf{PSNR↑} & \textbf{SSIM↑} & \textbf{LPIPS↓} & \textbf{CLIP↑} & \textbf{SigLIP↑} \\
\midrule
\multicolumn{6}{l}{\textbf{Text-to-3D}} \\
MVDream & \textbf{16.95} & 0.717 & 0.363 & 64.25 & 34.81 \\
MV-Adapter (TX) & 15.37 & 0.632 & 0.459 & 59.62 & 30.18 \\
SPAD & 8.34 & 0.619 & 0.468 & 61.32 & 29.12 \\
TRELLIS (TX) & 16.53 & \textbf{0.743} & \textbf{0.327} & 60.67 & 30.98 \\
\midrule
\multicolumn{6}{l}{\textbf{Image-to-3D}} \\
ImageDream-P & 15.50 & 0.728 & 0.400 & 60.89 & 31.67 \\
ImageDream-L & 15.64 & 0.732 & 0.393 & 61.72 & 32.52 \\
MV-Adapter (IM) & 15.24 & 0.646 & 0.448 & 61.46 & 32.00 \\
Era3D & 12.44 & 0.722 & 0.378 & 58.79 & 29.94 \\
TRELLIS (IM) & 16.02 & 0.741 & 0.378 & 55.39 & 26.72 \\
\midrule
\multicolumn{6}{l}{\textbf{3D Personalization}} \\
MVDreamBooth & 16.31 & 0.716 & 0.381 & 61.68 & 32.14 \\
\midrule
MV-RAG (Ours) & 16.63 & 0.730 & 0.362 & \textbf{64.48} & \textbf{35.34} \\
\bottomrule
\end{tabular}
\vspace{-0.3cm}
\label{tab:ind_compare_full}
\end{wraptable}

%% file: figures/ablation_k.tex
\begin{wrapfigure}[12]{r}{0.4\textwidth}
    \centering
    \vspace{-0.4cm}
    \includegraphics[width=1.0\linewidth]{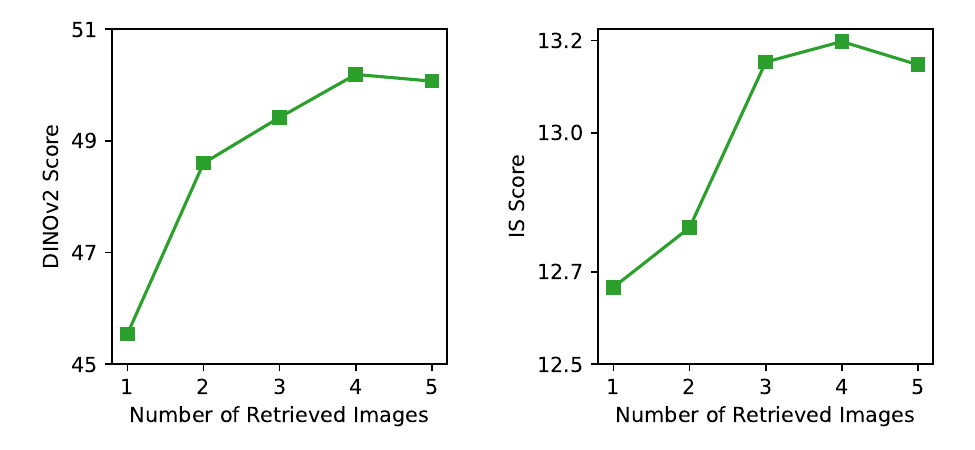}
    \caption{Effect of the number of retrieved images on alignment and fidelity. }
    \label{fig:ablation_k}
    
\end{wrapfigure}

%% file: figures/qualitative_results.tex
\begin{figure}[h]
  \centering
  \includegraphics[trim={0 0.1cm 0 0},clip,width=1\linewidth]{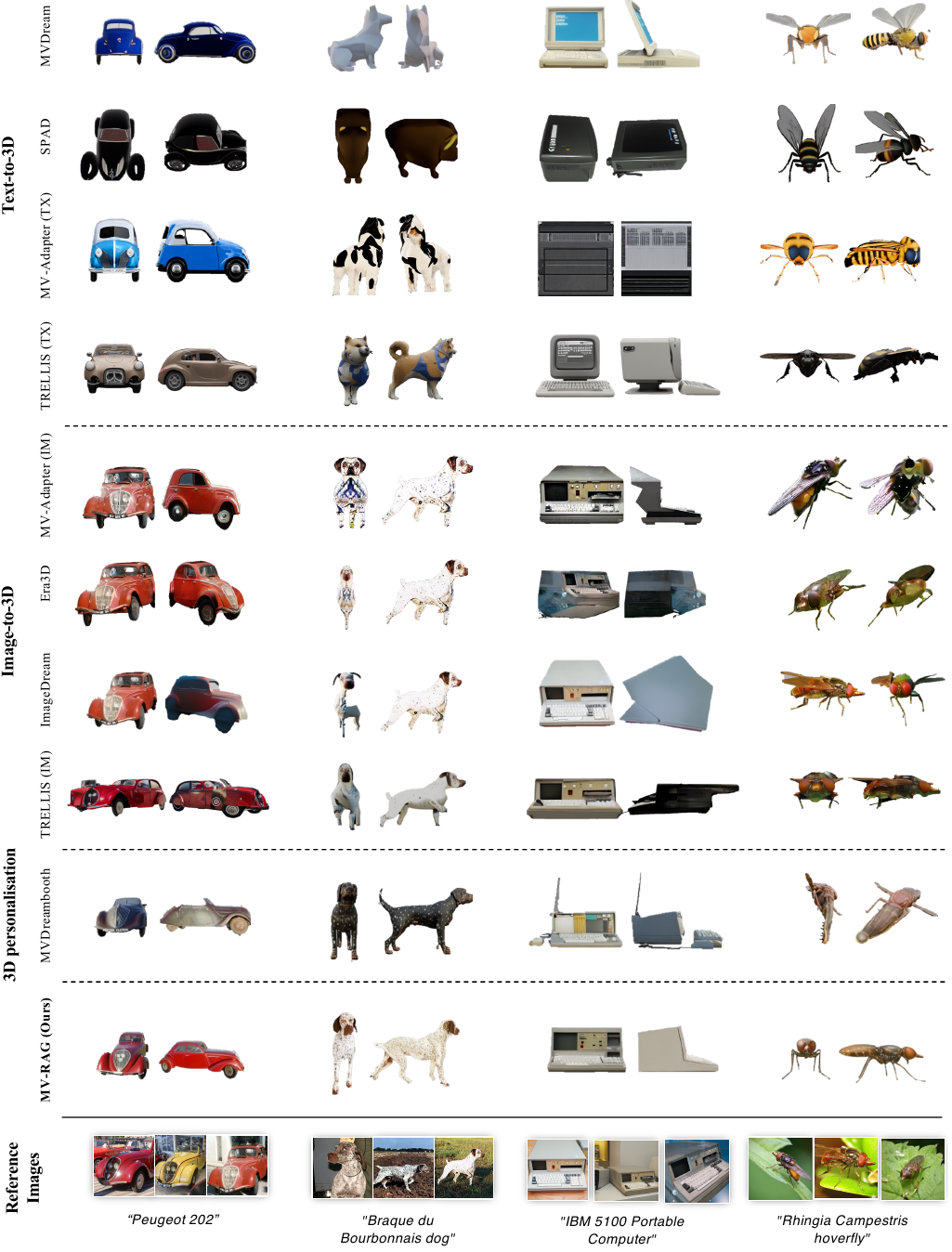}
    \caption{\textbf{Qualitative evaluation.} Text-to-3D models fail to generate unseen (OOD) concepts, while image-to-3D models fail to reconstruct their correct 3D structure from a single view. Existing personalization methods cannot effectively leverage the diversity of retrieved images.
    }  \label{fig:qualitative}
    \vspace{-0.7cm}
\end{figure}

%% file: tables/retrieval_k.tex
\begin{wraptable}{r}{5.3cm} 
\vspace{-0.7cm}
\centering
\caption{Comparison of retrieval approaches for out-of-domain retrieval.}
\begin{tabular}{lc}
\toprule
\textbf{Method} & \textbf{Precision@5} $\uparrow$ \\
\midrule
CLIP (TX-IM)   & 0.5366 \\
CLIP (TX-TX)   & 0.7306 \\
SigLIP (TX-IM) & 0.7889 \\
BM25 (TX-TX)   & \textbf{0.8522} \\
\bottomrule
\end{tabular}
\label{tab:retrieval_k}
\vspace{0.2cm}
\end{wraptable}

%% file: sections/06_conclusion.tex
\section{Conclusion}
\label{sec:conclusion}
We present a novel retrieval-augmented multiview diffusion model for text-to-3D generation. By retrieving relevant 2D images from a large-scale database and conditioning a multiview diffusion model on these images, our approach produces consistent and accurate multiview outputs—especially for out-of-domain (OOD) or rare concepts where prior methods often struggle. Central to our approach is a hybrid training scheme that effectively integrates structured multiview data with diverse 2D image collections, leveraging augmented conditioning views and a unique held-out view prediction objective. To thoroughly assess performance on challenging scenarios, we introduce a new benchmark of OOD prompts. Experimental results demonstrate that our method significantly improves 3D consistency, photorealism, and text alignment for OOD and rare concepts, while maintaining strong performance on standard benchmarks.

%% file: sections/07_appendix.tex
\newpage
\appendix
\section{Appendix}
\label{sec:appendix}

\section{Additional Qualitative Evaluation}

We present additional qualitative evaluations, corresponding to Fig.~4 of the main paper in Fig.~\ref{fig:comp_res} and Fig.~\ref{fig:comp_res2}. Additional examples for output of our method are provided in Fig.~\ref{fig:additional1} and Fig.~\ref{fig:additional2}. 

\section{Training, Inference and Implementation Details}
\label{sec:supp_implementation_details}
\subsection{Data Preparation} For 3D mode training, we utilize multi-view synthetic renderings from the public Objaverse dataset \cite{qiu2024richdreamer, deitke2023objaverse} along with the associated camera parameters. We randomly sampled a subset of 90K objects. For each object, we select four orthogonal views with elevation angles in the range $[-5^\circ, 30^\circ]$ for supervision. Additionally, we sample 2-3 random views to simulate retrieval, as detailed below. Objaverse contains a wide variety of objects, including both high-fidelity, photorealistic assets and low-textured, abstract ones. To improve the model's robustness to real-world, non-synthetic data, we apply an aesthetic-based filtering criterion. This criterion incorporates color diversity, texture complexity, and multi-view consistency, which then results in about 65K objects. Following the preprocessing protocol of MVDream, we resize all rendered images to $256 \times 256$ pixels and replace empty backgrounds with a random gray color. Camera poses are normalized onto a unit sphere by removing translational components.

To simulate retrieval images, we apply a series of augmentations to the additional rendered views. These include perspective distortion, random rotations, resized cropping, and color jitter. 
To further enhance realism, we employ an image-variation model~\cite{Ye_etal_2023_IPAdapter, rombach2022highresolutionimagesynthesislatent} to generate semantically and visually diverse variants of the same object. In total, we obtain four simulated retrieval images per object.
For 2D mode training, we use the ImageNet21K dataset\cite{ridnik2021imagenet}, which comprises over 21K semantic classes with multiple images per class. To improve the visual coherence within classes, we use a large language model (GPT-4o) to filter and retain only visually unified categories (e.g., \textit{carpet shark}, \textit{toilet bowl}) and exclude abstract or overly broad classes (e.g., \textit{human}, \textit{cycling}). This results in a curated subset of 516 visually consistent categories. For each selected class, we sample one target image for supervision and four additional images from the same class to serve as retrieved images.

\subsection{Training} We fine-tune our model using the AdamW optimizer\cite{loshchilov2019decoupledweightdecayregularization} with a learning rate of $5 \times 10^{-6}$ and a batch size of 24 for approximately 11{,}000 steps. Training is performed in an alternating scheme between 2D and 3D modes, allocating an equal number of steps to each mode. As in MVDream, we append ", 3d asset" to the text prompt during 3D mode to help the model distinguish between the two training regimes.
The model is initialized from the Stable Diffusion 2.1-based MVDream checkpoint, which remains frozen throughout training. The adapter modules are initialized from the ImageDream checkpoint. We fine-tune both the retrieval-attention modules and the Resampler. For the image encoder, we use OpenCLIP ViT-H/14, which is kept frozen during training.
The training was done on a single NVIDIA A100 GPU, with a total training time of approximately 3 hours.

\subsection{Baselines}
\label{sec:supp_baselines}
For all baselines we use the official implementations and publicly available pretrained checkpoints provided by the respective authors, with the exception of MVDreamBooth, for which training code is not released.
For each baseline, we generate 4 views using fixed orthogonal camera angles and elevations, employing the DDIM sampler with 50 steps and a classifier-free guidance (CFG) scale of 5.
For image-to-3D baselines, we preprocess the retrieved reference images by segmenting out the background using Grounded-SAM~\cite{kirillov2023segany, liu2023grounding, ren2024grounded}. Among the reference images, we select the one that yields the highest multi-view consistency based on DINO score for evaluation. To ensure a fair comparison in image-image similarity metrics, we compare semantic features against the segmented ground-truth object views.

\input{figures/clip_issues}
For the MVDreambooth baseline, we follow the method described in~\cite{Shi_etal_2023_MVDream}, training a separate MVDreamBooth model for each instance in the OOD-Eval set. Each model is optimized for 600 steps. To preserve class identity, we apply a class-preservation loss using ImageNet class names (e.g., \textit{dog}, \textit{car}) when available, and default to the prompt when no corresponding class is derived.
\input{figures/diversity_illustration}
\subsection{Metrics}
\label{sec:supp_metrics}
Figure~\ref{fig:clip_issues} highlights three representative failure cases of using CLIP text-image similarity as an evaluation metric for out-of-distribution (OOD) objects. In each case, MVDream receives a higher CLIP score than both our model and even the ground truth image, despite generating outputs that are visually or semantically incorrect. We hypothesize that this stems from CLIP's limited prior knowledge of rare concepts and the fact that models like MVDream are optimized to align with CLIP-based features, potentially leading to overfitting to incorrect semantic associations. Further, as shown in Table~\ref{tab:clip-retrieval-scores}, we find that for rare concepts, CLIP assigns nearly identical similarity scores to both detailed object names and their coarse class labels. This suggests that CLIP does not treat the additional semantic information in rare object names as meaningful, highlighting a lack of conceptual grounding for these OOD categories. In contrast, for in-domain objects, CLIP shows much stronger separation between specific and generic labels, reinforcing its limitations in recognizing and evaluating uncommon or unseen concepts.

These observations underscore the limitations of using CLIP for OOD evaluation and motivate our decision to adopt image-image similarity metrics instead, which more reliably reflect visual fidelity. To this end, we employ CLIP~\cite{Radford_etal_2021_CLIP}, DINOv2~\cite{oquab2023dinov2}, and an Instance Retrieval (IR) model~\cite{shao20221st} fine-tuned from CLIP to better align visual object instances.

\input{figures/benchmarks}
\subsection{Re-rendering}
To more thoroughly assess the 3D consistency and fidelity of the baselines on the OOD-Eval benchmark, we employ LGM~\cite{tang2024lgm}. Specifically, we reconstruct a 3D Gaussian representation using four output views generated by each model. From this reconstruction, we render 18 additional novel views sampled along a circular trajectory. These re-rendered views are then evaluated against the retrieved images using the image-image similarity metrics described earlier. We utilize the publicly available LGM implementation with its default configuration settings.

\input{tables/clip_fails_rare}
\subsection{Retrieval Process} \label{subsec:retrieval_process}
We evaluate multiple retrieval strategies based on both image-text and text-text similarity. For embedding-based retrieval with CLIP~\cite{Radford_etal_2021_CLIP}and SigLIP~\cite{zhai2023sigmoid}, we build an index using the FAISS library~\cite{douze2025faisslibrary}, which supports efficient approximate nearest-neighbor search in high-dimensional spaces. We additionally employ Pyserini~\cite{Trotman2014ImprovementsTB,Robertson1994OkapiAT,Lin_etal_SIGIR2021_Pyserini} for text-based retrieval using a Bag-of-Words approach powered by the BM25 ranking function. This approach is a highly optimized and scalable toolkit designed for large-scale retrieval tasks, capable of indexing millions of documents while providing fast query responses. Its retrieval time is typically sub-linear with respect to the size of the corpus due to inverted index structures, enabling near real-time search performance with minimal computational overhead, as demonstrated in large-scale search engine systems.

\section{User Study}
\label{sec:supp_user_study}
We provide additional details about the user study referenced in Tab.\ref{tab:user_study}. The study involved 8 different objects, each evaluated using 3 methods: MV-RAG, MVDream, and ImageDream. For each object, participants were first shown a brief text description along with two sample images of the object to establish context. They were then shown sets of four images corresponding to different views-generated by each method. The internal order of the methods was randomized per object to mitigate ordering bias. Participants were asked to rate the following three questions on a scale from 1 to 5: (Q1) “How well do the 4 images match {object}?“, (Q2) “How realistic do the 4 images look overall?“, and (Q3) “How well do the 4 images appear to be consistent with each other, as if they show different views of the same 3D object?“. The study was conducted using Google Forms, and participants viewed the images on a computer screen. The user population consisted of 30 randomly selected individuals across diverse ages, ethnicities, and genders.

\section{Evaluation Dataset Construction}
\label{sec:benchmarks}
\paragraph{Construction of OOD-Eval.}
We construct an evaluation benchmark, OOD-Eval, consisting of 196 objects. To ensure diversity and out-of-distribution coverage, we use a large language model (GPT-4o) to curate object names representing rare or unique concepts, as well as familiar objects that are absent from the training data. These include examples such as extinct or rare animal species, uncommon vehicles, and other atypical items. A visual preview of the benchmark is provided in Fig.~\ref{fig:benchmarks}.

For generating image captions, we leverage a vision-language model, specifically Qwen-VL~\cite{bai2023qwenvlversatilevisionlanguagemodel} which provides high-quality textual descriptions of the images. These captions are used in the retrieval process (see Sec~\ref{sec:inference}).

\paragraph{Construction of IND-Eval.}
We constructed an in-domain evaluation set by selecting 50 well-known or everyday objects from the widely used Objaverse-XL dataset. For each object, we retrieve 4 reference images from the large-scale LAION-400M dataset~\cite{schuhmann2021laion} using BM25-based text retrieval (see Sec.3 in main paper).  The retrieved images often exhibit significant visual or modality variation (e.g., artistic renderings or paintings of the object), as illustrated in Fig.~\ref{fig:benchmarks}

\input{figures/limitations}
\section{Limitations}
While effective, our method has several limitations. It relies heavily on both the quality of the retrieved image corpus and the capabilities of the underlying generative model, MVDream. When the base model lacks prior knowledge of the object and retrieval fails to provide informative or diverse references, the generated multiviews can be inaccurate or implausible.

As shown in Fig.~\ref{fig:limitations}, errors may arise when the retrieved images are visually biased-e.g., all showing similar white flowers, leading to reduced diversity and visual artifacts. Furthermore, our training objective promotes texture variation, which can make it difficult to reproduce fine-grained or specific patterns, such as the hoverfly's dorsal markings.

Our model also employs an adaptive mechanism that balances attention between the base model and the retrieval adapters, based on a similarity score between the generated initial views and retrieved images. When the base model demonstrates high similarity to the target object but exhibits 3D structural errors (such as a floating dog tail), these artifacts may be inherited. This limitation could be addressed by incorporating a more sophisticated and 3D aware scoring function.

Lastly, our method introduces a retrieval phase prior to generation. Although this adds computational cost relative to standard text-to-image pipelines, the overhead is minimal. As shown in Section~\ref{subsec:retrieval_process}

\section{Broader Impact}
Our model advances multiview image generation by conditioning on text prompts and retrieved image corpora, enabling applications in graphics, virtual reality, and content creation. Positively, this technology can facilitate more immersive and accurate 3D visualizations and assist artists or designers in generating diverse object views with limited data. However, improved generative capabilities also pose risks, such as enabling the creation of highly realistic synthetic images that could be misused for disinformation or deepfakes. Additionally, biases present in the retrieval corpus or base models may propagate or amplify undesirable stereotypes or inaccuracies in generated outputs. To mitigate such risks, future deployment should include careful curation of training and retrieval datasets, transparency about generated content, and potentially gating access to the most advanced models.
\input{figures/additional_results}
\input{figures/results_comparison}

%% file: figures/clip_issues.tex
\begin{wrapfigure}[30]{r}{0.40\textwidth}
  \centering
  \vspace{-0.3cm}
  \includegraphics[width=\linewidth]{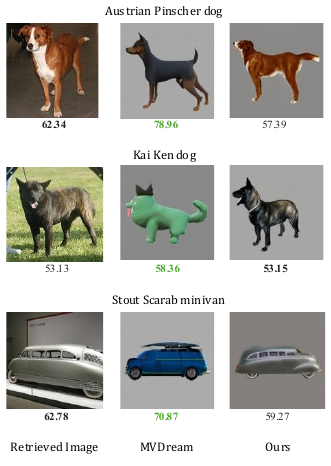}
  \caption{\textbf{Limitations of CLIP text-image similarity for evaluating OOD objects.} 
  Each row shows an example from our OOD-Eval benchmark: the ground truth (GT) image, the output from MVDream, and the output from our model. Below each image is its CLIP similarity score. MVDream receives a higher score than both our model and the GT image, despite producing less faithful generations.
  }
  \label{fig:clip_issues}
\end{wrapfigure}

%% file: figures/diversity_illustration.tex
\begin{figure}[t]
  \centering
  \vspace{-0.5cm}
  \begin{tabular}{cc}
    \includegraphics[trim={0cm 0 0 0},clip,width=0.47\linewidth]{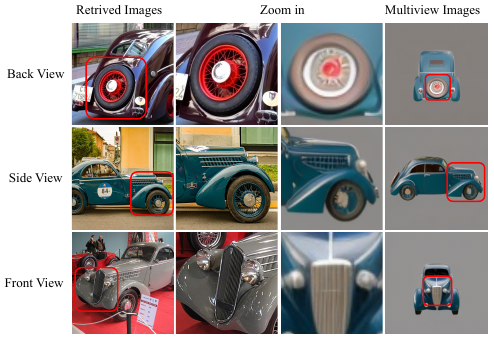} & 
    \includegraphics[trim={0 0.05cm 0 0cm},width=0.43\linewidth]{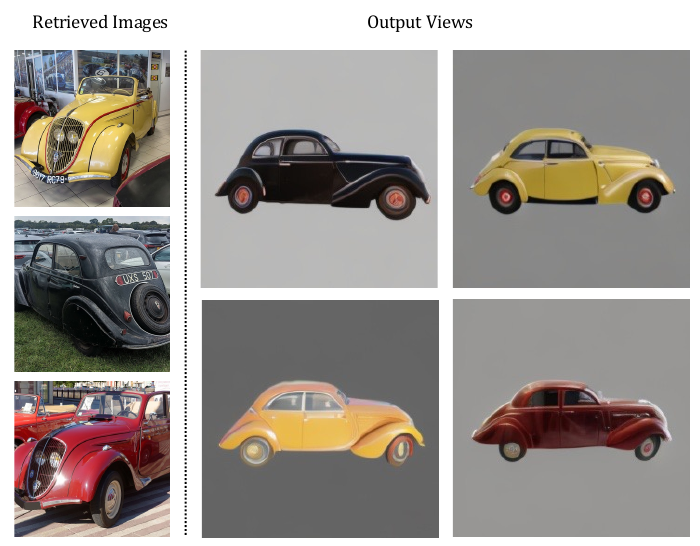} \\
    (a) & (b)
  \end{tabular}
  \caption{(a). \textbf{Utility.} Our approach learns to utilize all relevant information in retrieved views. On the LHS, we show retrieved views. The middle columns are zoom ins, for aspects used in generation, and the RHS shows back view (top), side view (middle), and front view (bottom). (b). \textbf{Diversity.} Unlike image-prompted methods, our method enables diverse outputs, by using different seeds for a given input text and a fixed set of retrieved images. For the prompt "Peugeot 202", the LHS shows retrieved views, and the RHS shows a single view output (using the same pose) for 4 different seeds.}
  \label{fig:diversity}
  \vspace{-0.5cm}
\end{figure}

%% file: figures/benchmarks.tex
\begin{figure}[H]
  \centering
  \includegraphics[width=\linewidth]{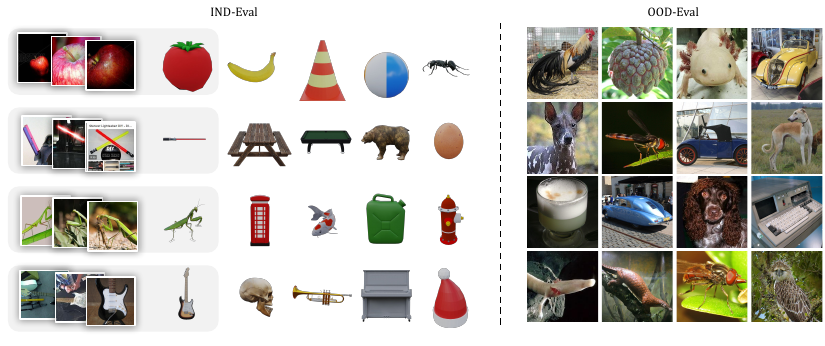}
  \caption{\textbf{Evaluation Benchmarks Overview.} 
\textbf{OOD-Eval:} Our out-of-distribution benchmark includes 2D images of both rare and well-known objects, featuring a diverse set of categories such as animals, vehicles, insects, foods, and everyday items. 
\textbf{IND-Eval:} The in-domain benchmark focuses on common, everyday objects that are representative of standard training distributions.} 
  \label{fig:benchmarks}
\end{figure}

%% file: tables/clip_fails_rare.tex
\begin{wraptable}{r}{0.5\textwidth} 
\vspace{-8pt} 
\caption{\textbf{CLIP similarity between retrieved images and concept labels.} We compute similarity to the GT retrieved images with the object name against the class label (e.g., "dog", "car") and report average, max, and their absolute difference. OOD examples show minimal semantic separation.}
\centering
\scriptsize  
\renewcommand{\arraystretch}{1.1}
\begin{tabular}{l|l|c|c}
\toprule
\textbf{Domain} & \textbf{Text} & \textbf{Avg} & \textbf{Max} \\
\midrule
\multirow{3}{*}{OOD}
& Bucovina Shepherd Dog & 63.59 & 67.16 \\
& Dog & 63.41 & 67.45 \\
& \textit{Abs. Diff} & \textit{\textbf{0.18}} & \textit{\textbf{0.29}} \\
\cmidrule{2-4}
& BMW 319 automobile car & 66.33 & 71.11 \\
& Car & 65.14 & 71.56 \\
& \textit{Abs. Diff} & \textit{\textbf{1.19}} & \textit{\textbf{0.45}} \\
\midrule
\multirow{3}{*}{In-Domain}
& Airedale Terrier dog & 89.54 & 96.58 \\
& Dog & 64.23 & 69.99 \\
& \textit{Abs. Diff} & \textit{25.31} & \textit{26.59} \\
\cmidrule{2-4}
& American Hairless Terrier dog & 83.08 & 92.79 \\
& Dog & 64.59 & 70.11 \\
& \textit{Abs. Diff} & \textit{18.49} & \textit{22.68} \\
\bottomrule
\end{tabular}
\label{tab:clip-retrieval-scores}
\end{wraptable}

%% file: figures/limitations.tex
\begin{wrapfigure}[28]{r}{0.45\textwidth} 
  \centering
  \vspace{-0.5cm} 
  \includegraphics[width=0.85\linewidth]{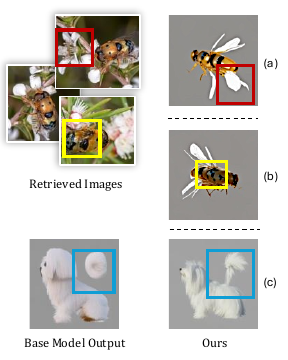}  
  \vspace{-0.4cm}  
  \caption{\textbf{Limitations.} (a) Visually biased retrieved images (e.g., repetitive white flowers) introduce artifacts in the generated multiviews.
(b) The model struggles to reproduce fine-grained textures, such as the hoverfly’s dorsal pattern.
(c) When the base model (MVDream) is assigned a high attention weight ($\alpha$), 3D structural inaccuracies from the base model (e.g., a floating tail) are inherited.}
  \label{fig:limitations}
  \vspace{-0.3cm}  
\end{wrapfigure}

%% file: figures/additional_results.tex
\begin{figure}[h]
  \centering
  \includegraphics[trim={0 0cm 0 0},clip,width=1\linewidth]{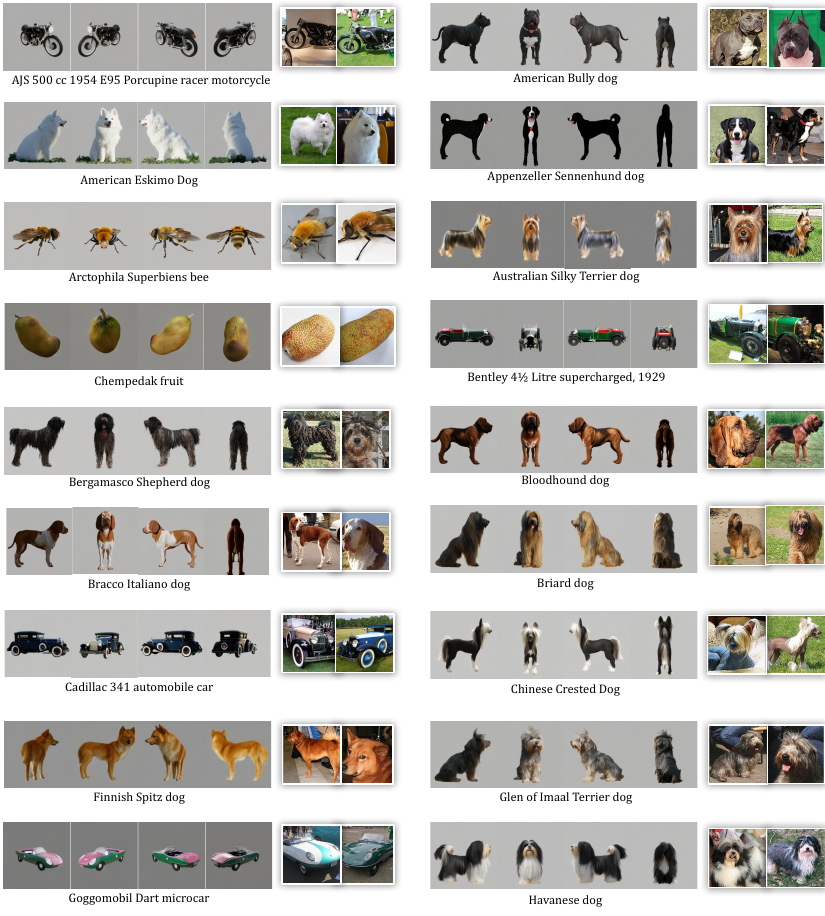}
\caption{\textbf{Additional Results.}  
    }
    \label{fig:additional1}
    \vspace{-0.4cm}
\end{figure}

\begin{figure}[h]
  \centering
  \includegraphics[trim={0 0cm 0 0},clip,width=1\linewidth]{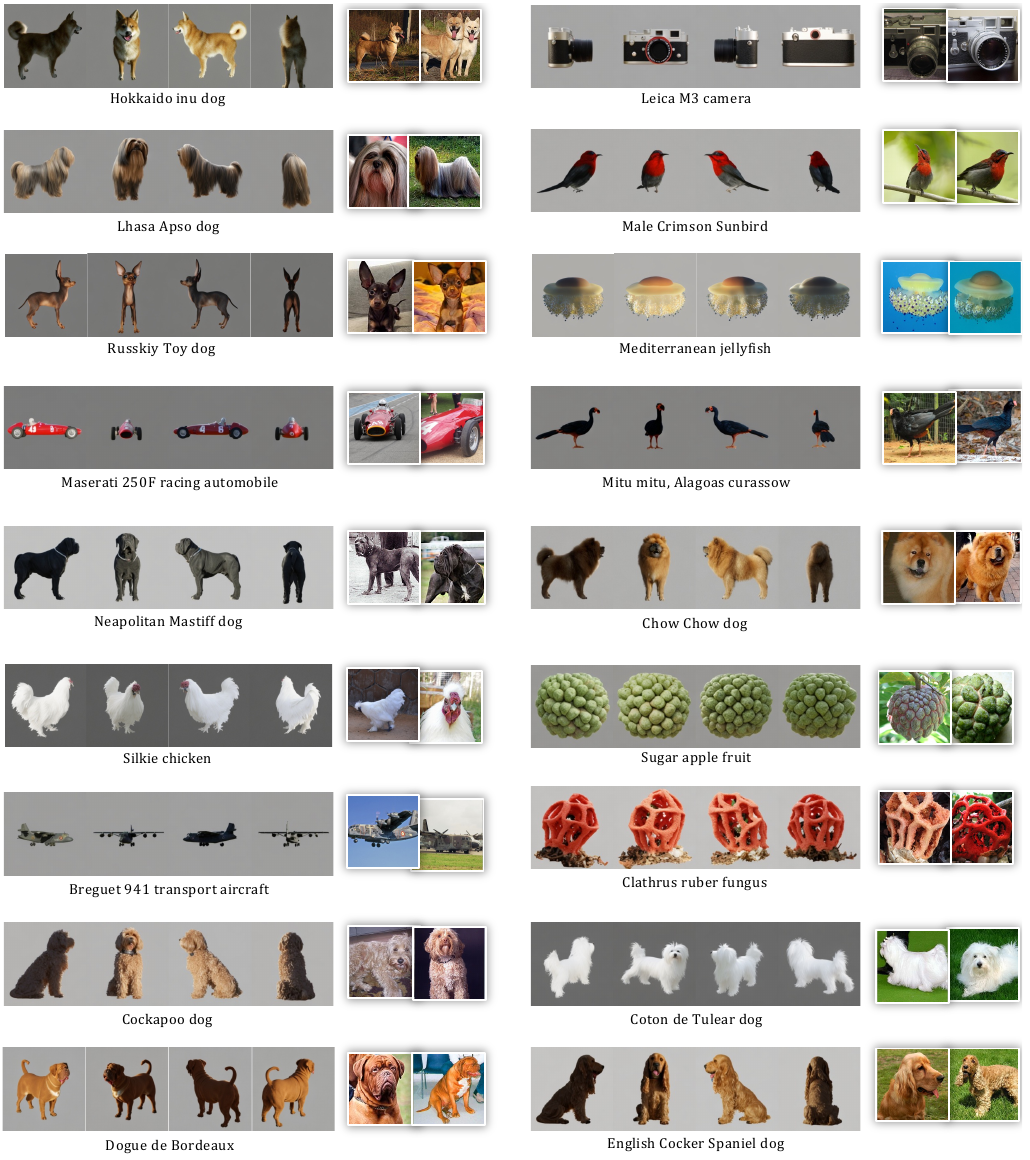}
    \caption{\textbf{Additional Results.}
    }  \label{fig:additional2}
    \vspace{-0.4cm}
\end{figure}

%% file: figures/results_comparison.tex
\begin{figure}[t]
  \centering
  \includegraphics[trim={0 0.3cm 0 0},clip,width=1\linewidth]{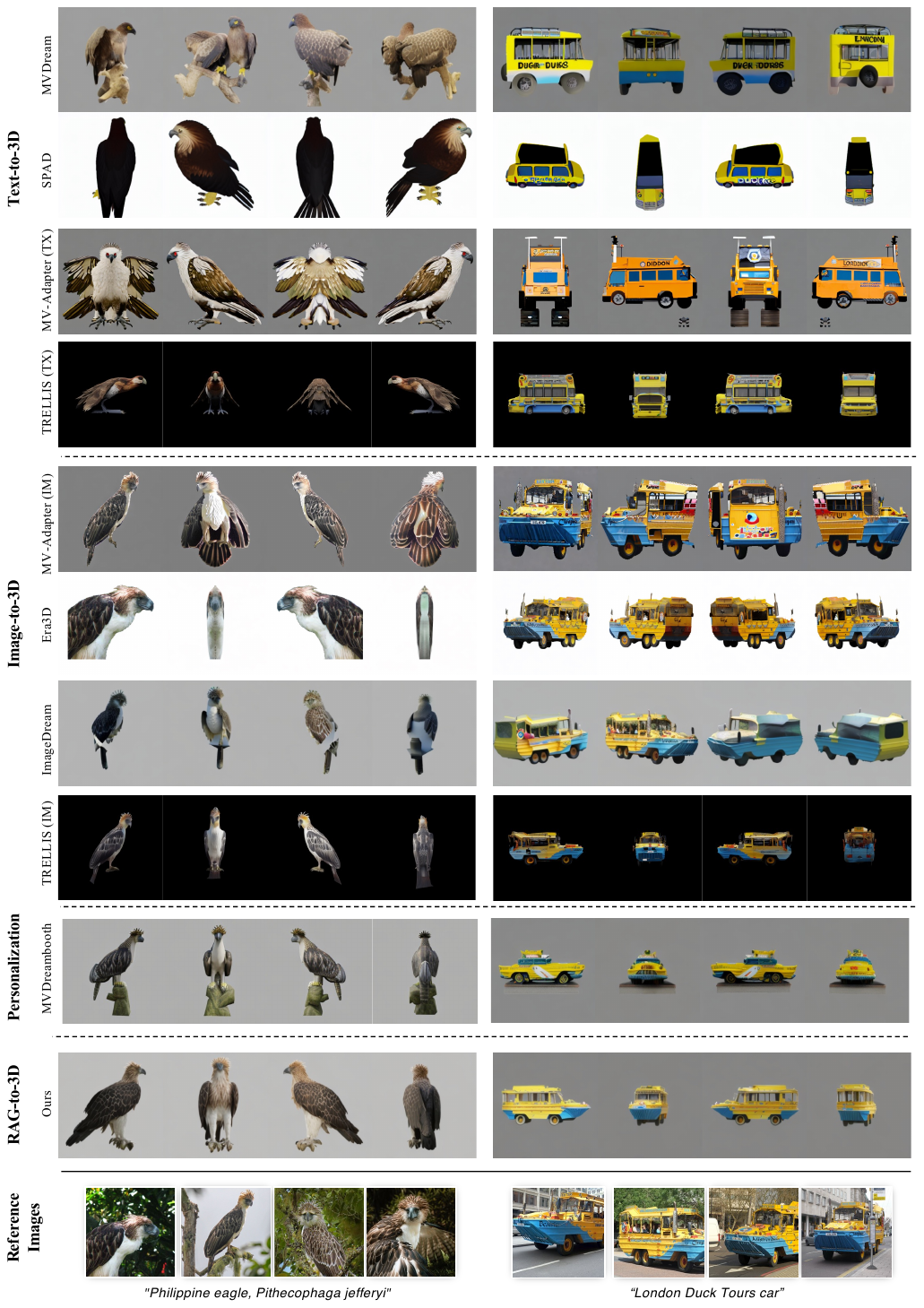}
  \caption{\textbf{Additional qualitative evaluation.} Additional examples to those shown in Fig.~\ref{fig:qualitative}.
    }
    \label{fig:comp_res}
\end{figure}

\begin{figure}[t]
  \centering
  \includegraphics[trim={0 0.2cm 0 0},clip,width=1\linewidth]{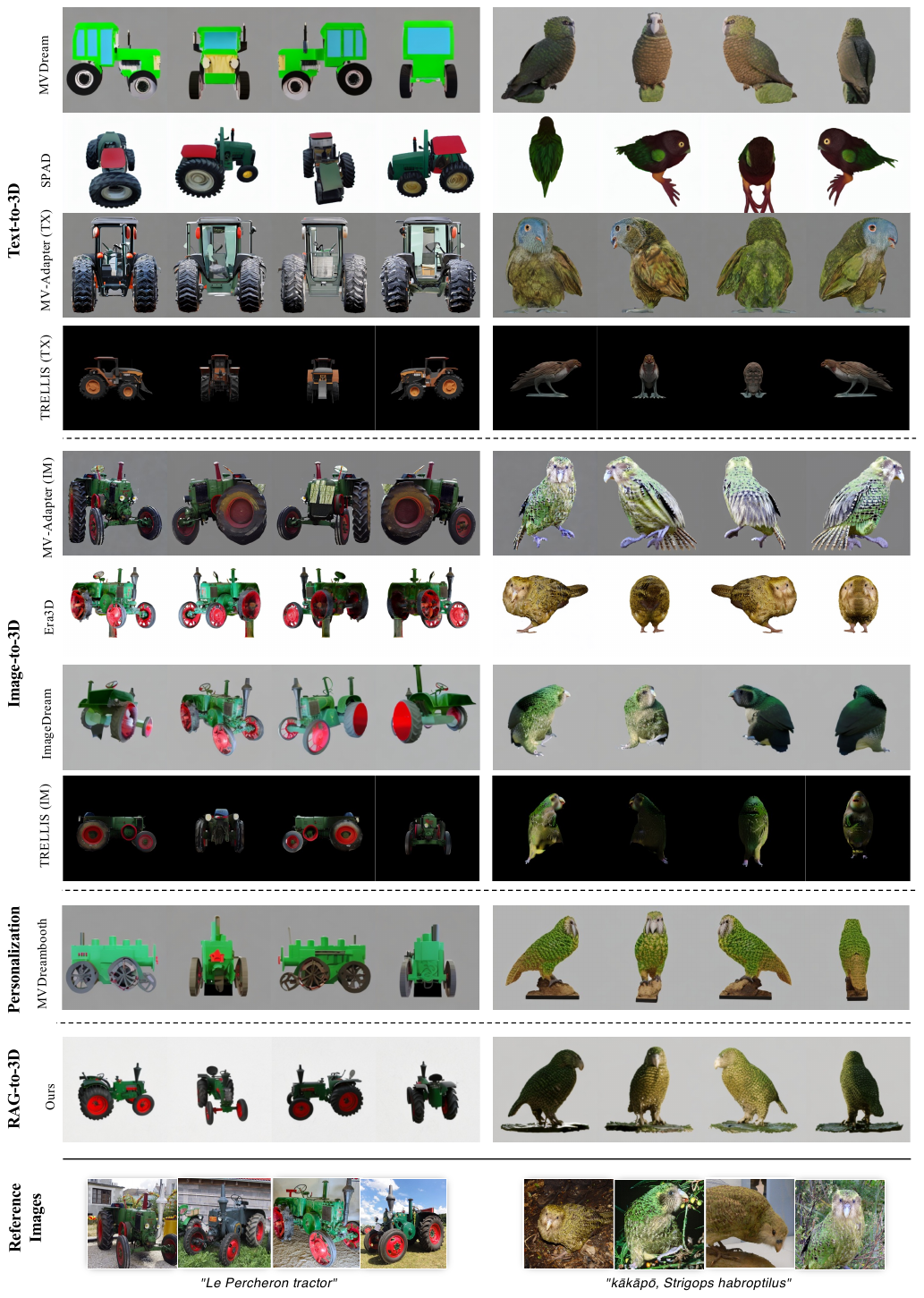}
    \caption{\textbf{Additional qualitative evaluation.} Additional examples to those shown in Fig.~\ref{fig:qualitative}.
    }  \label{fig:comp_res2}
    
\end{figure}